# Using Recursive K-Means and Dijkstra Algorithm to Solve CVRP

Author: Hassan Moussa

*Lebanese University, Department of applied mathematics*

**Abstract:**

Capacitated vehicle routing problem (CVRP) is being one of the most common optimization problems in our days, considering the wide usage of routing algorithms in multiple fields such as transportation domain, food delivery, network routing, ... Capacitated vehicle routing problem is classified as an NP-Hard problem, hence normal optimization algorithm can't solve it.

In our paper, we discuss a new way to solve the mentioned problem, using a recursive approach of the most known clustering algorithm "K-Means", one of the known shortest path algorithm "Dijkstra", and some mathematical operations. In this paper, we will show how to implement those methods together in order to get the nearest solution of the optimal route, since research and development are still on go, this research paper may be extended with another one, that will involve the implementational results of this thoric side.

## I. Introduction:

Capacitated vehicle routing problem is finding the optimal route for a fleet of vehicles to pass through a number of clients, given a set of constraints. The CSVP was very extremely studied in the field of optimization, because of its wide uses, and the whole evolution in the E-commerce and delivery field, also the spread of school and taxi operators.

Capacitated vehicle routing problem is a subset of vehicle routing problem (VRP) which is very common, CVRP is related to the size of vehicle, which is considered as a constraint. In other words, VRP is the problem of finding the least cost of routes from a depot to a set of geographical points.

In order to define the capacitated vehicle routing problem in mathematical terms we will start by defining some parameters:

**n** is the number of clients

- N is set of clients, with N= {1,2, ..., n}
- V is set of nodes with V = {0} ∪ N
- A is a set of arcs, with $A = \{ (i,j) \in V^2 : i \neq j \}$
- $C_{ij}$ is the cost of travel over arc $(i,j) \in A$
- Q is the vehicle capacity

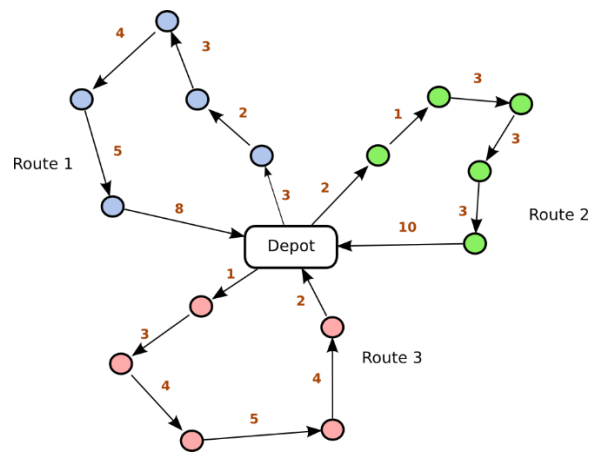

*Figure 1: Graph of CVRP / https://neo.lcc.uma.es/dynamic/images/vrp.png*

The formulation in terms of mathematical operation will be:

$$\min_{i,j \in A} \sum_{i,j \in A} c_{ij} \, x_{ij}$$

Which mean that we need to minimize the whole cost.

$$\sum_{j \in V, j \neq i} x_{ij} = 1 \; while \; i \in N$$

if x is a client then it should belong to one node, and go only to one node.

$$\sum_{i \in V, i \neq j} x_{ij} = 1 \; while \; i \in N$$

If I reach a client, then I should come from another node.

## II. Method:

The CVRP problem can be divided into multiple sub-problems, in our paper we will follow the following approach:

1. Finding the optimal distribution of students along the vehicles, taking into consideration their capacity, this will help the algorithm in optimizing the overall occupacity
2. Clustering the clients into a number of clusters that suits the number of vehicles and their capacity using our recursive approach of "K-Means" algorithm, and then assign each cluster to a vehicle, taking into consideration the occupacity of the vehicle, the ride time, and the ride distance.
3. Performing a second phase of optimization, this phase will make sure that the occupacity of a vehicle is optimal taking into consideration the ride distance and time, in case any of our constraints is touched, we go through a hierarchical clustering to join or split a specific cluster.
4. Performing the routing between each node of the cluster and the most near one, and then between the depot ({0}) and the nearest node of a cluster, in our approach, we are using an open shortest path algorithm "Dijkstra", we represent the considered city or country as a mathematical graph, where every location is a vertex, and the existence of a route between any 2 vertices is considered as an edge, using "Dijkstra" we can find the shortest path to travel through all nodes.

### A. Optimal distribution of students in vehicles:

Vehicles in Capacitated Vehicle Routing problem can have a specific capacity, each vehicle can differ from others, which will be directly related to an important insight, the "Occupacity", one of our algorithm KPI, so in order to build the perfect solution we need to start by finding the best distribution of clients.

Let's start by defining the problem in mathematical terms, let's consider that we have t number of vehicles, and n the number of clients.

$$V = \{v1, v2, ..., vt\} \text{ set of } t \text{ vehicles}$$

$$C = \{c1, c2, ..., ct\} \text{ set of assigned capacity}$$

$$ci = Capacity(vi) \text{ for } 1 \leq i \leq t$$

$$x \in R^n, x = \begin{pmatrix} x_1 \\ \vdots \\ x_t \end{pmatrix}$$

$$x_i = \text{number of } v_i \text{ needed}, 0 \leq i \leq t$$

The final equation to find the optimal number of each vehicle needed is:

$$f(x) = n$$

Known that:

$$f(x) = \sum_{i=1}^{t} x_i c_i = x_1 c_1 + x_2 c_2 + \cdots + x_t c_t$$

f(x) is a defined function $R^n \to R$, which find the number of each vehicle needed x, taking into consideration that the sum of the combination between the capacity and number of vehicles assigned should equal to the total number of clients.

In order to find x, we are going to use one of known method in numerical analysis, "Newton-Raphson Method" applied to n variables.

**Objective**: find an approximation of the equation $f(x) = n \to f(x) - n = 0$

**Input**:
- $F : R^n \to R$
- $Jacobiane\ Matrice\ J_f : R^n \to R^n$
- A first approximation $X_0 \in R^n$
- The precision required $\varepsilon \in R, \varepsilon > 0$

**Output**:

An approximation of the solution $x^* \in R^n$

**Initialization**:

$$\alpha = 0$$

**Iterations**:

$$d_{\alpha+1} = \frac{-F(x_\alpha)}{J_f(x_\alpha)}$$

$$J_f(x_\alpha) = \left(\frac{\partial f}{\partial x_1}(x_\alpha), \frac{\partial f}{\partial x_2}(x_\alpha), \dots \frac{\partial f}{\partial x_t}(x_\alpha)\right)$$

**Set**: $x_{\alpha+1} = x_\alpha + d_{\alpha+1}$

**Set**: α = α + 1

**Stop When**: $\|F(x_\alpha)\| \leq \varepsilon \to x^* = x^\alpha$

$x^*$ is a distribution approximation of the clients and vehicles, so we need $x_1$ number vehicles $v_1$ of capacity $c_1$.

**B. Recursive Clustering "K-Means":**

Clustering in one of the most important phases of any routing algorithm, because it will affect directly the final results and KPIs of our algorithm, in our approach, we are clustering our clients in a recursive approach, we start by dividing the client's data in two clusters based on their geolocation, and then we apply the same "K-Means" again on the two clusters, until we reach the threshold, which will be one of the vehicle capacity, in this case, this cluster will be assigned to this vehicle.

Each iteration of the recursive "K-Means" is saved in a phase tree, the root phase will be the initial data, then the phase 2 and 3 will be the left and right nodes represented as clusters.

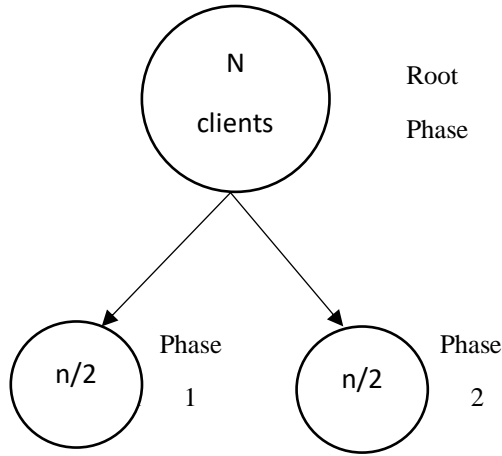

Let's start by defining the K-Means algorithm, the objective function of this algorithm can de represented in the following shape:

$$J = \sum_{j=1}^{k}\sum_{i=1}^{n}\left\|x^{(j)}{}_i - c_j\right\|^2$$

Where: k is number of clusters, n is number of cases, we will define the algorithm step by step in the following:

**Input**: k number of clusters, $x_1, x_2, \ldots, x_n$ n elements to be clustered

**Algorithm**:

Consider $c_1, c_2, \ldots, c_k$ k random centroid

**Repeat Until Done:**

**For each $x_i$:**

    1. Get nearest centroid $c_j$:

$$arg \min_j D(x_i, c_j)$$

    2. Assign $x_i$ to $c_j$

    3. for each cluster j = 1,2, ..., k

        Get the new center:

$$center = \frac{1}{n_j}\sum x_i D(x_i, c_j)$$

The algorithm will first set a random centroid, assign nodes to the nearest one, then calculate the new center, and assign nodes to new centroid. Iterating over this operation until nodes are assigned to the same centroid.

Because we are dealing with geolocations, we are using haversine as a distance function, which can be defined as the following:

$$d = 2r \arcsin\left(\sqrt{\sin^2\left(\frac{\phi_2 - \phi_1}{2}\right) + \cos(\phi_1)\cos(\phi_2)\sin^2\left(\frac{\lambda_2 - \lambda_1}{2}\right)}\right)$$

*Figure 2: Haversine Formula, https://user-images.githubusercontent.com/2789198/27240436-e9a459da-52d4-11e7-8f84-f96d0b312859.png*

Regarding the time complexity of this algorithm, lets consider that our algorithm iterates n times, have k cluster, w nodes and d dimensions, so the complexity of the regular "K-Means" is:

$$O(n * k * w * d)$$

In our case, we will split each cluster into 2, and we are working on 2 dimensions, so the complexity will be for an iteration of recursive "K-means":

$$O(4 * n * w)$$

After having defined the regular "K-Means" clustering algorithm, we will go to implement our approach used in geolocation data, which is the recursive "K-Means"

**Input**:

Root Phase (initial dataset of geolocation)

**Algorithm**:

- $c_1, c_2 = Kmeans(root)$
- $root.right = c1$
- $root.left = c2$
- $KMeans(root.right)$
- $KMeans(root.left)$

**Stop Case**:

The number of nodes assigned to a cluster is equal to one of our vehicle capacities $x^*$, stop iterating through this cluster and assign it to the mentioned vehicle.

**Output**:

All leaves of the tree are the final result of our algorithm, they will be the clusters to be applied to routing algorithm later.

**C-Optimization Phase:**

After getting the final clusters by our recursive algorithm, some clusters may have a low number of assigned nodes, which may affect the over all occupancy rate, this is the reason of the optimization phase, it looks into all clusters, and see where we can edit to optimize the final KPIs, we take a parameter which is the minimum percentage of a cluster, if a cluster doesn't fit the minimum, we start by looking to other clusters who can handle the number of nodes available on it. This is done by calculating the distance between the centroid of the low occupancy cluster and the available others, and merge the nearest one.

This is a mathematical operation to get the following value:

$$\min_j \|D(c_i - c_j)\|$$

Which is the minimum distance between the centroid i of the affected cluster and all other available cluster.

Note: a cluster is considered available if:

$n_j + n_i < c_j$ which means that the nodes available on cluster j added to the one on cluster i are still under the over all capacity of the vehicle.

After the optimization phase, the most important part takes the initiative, "Routing Phase" which will take the responsibility of finding the optimal routing in a cluster.

**D- Routing Phase using "Dijkstra":**

We know that in the geolocation world, there is an infinity of route to go from a geolocation to another, and because we are dealing with vehicles, we restrict our research on the optimal driving route between two location. The recent output was an optimized cluster that involve nearest nodes, now we need to set a flow to know where to start and where to finish, and taking into consideration that the covering route for all nodes should be optimal.

We will start by considering the concerned city or country as a graph, where each location is a vertex, and a driving route between two location is considered as an edge, lets start by defining this graph:

$V = \{v1, v2, \ldots, vn\}$ set of vertices (locations)

$E \subseteq \{\{x, y\} \, x, y \in V \text{ and } x \neq y\}$

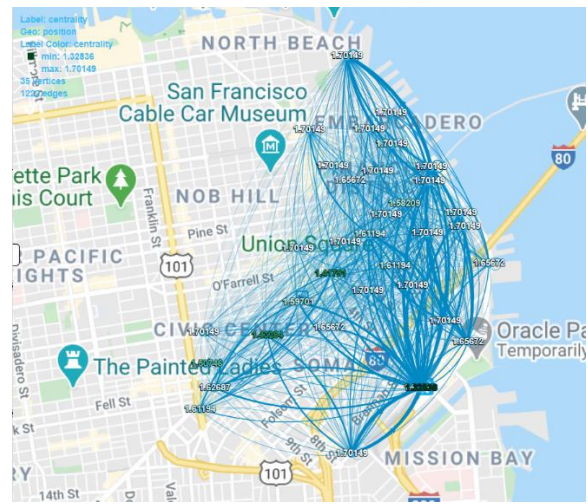

*Figure 3 Example of a graph applied on geo location data*
*https://miro.medium.com/max/1766/0*18KbdJeKuUBYm8JR*

Applying "Dijkstra" in the given graph:

**Input**: Graph G and the source node

**Algorithm**:

Set: $dist[source] = 0$

$Consider \, Q \, as \, a \, priority \, queue$

**For each** vertex v in Graph:

    If v ≠ source:

        $dist[v] = \infty$

        $pre[v] = undefined$

        $Q \leftarrow (v, dist(v))$

**While** Q is not empty:

    $u \leftarrow \min(Q)$

    For each neighbor v of u:

        $alt \leftarrow dist[u] + len(u,v)$

        If $alt < dist[v]$ :

            $dist[v] \leftarrow alt$

            $prev[v] \leftarrow u$

**Output**:

Distance and previous.

**So, the routing algorithm will be defined as follow:**

**Input**: clusters

**Algorithm**:

**For each** cluster c:

    Add cluster nodes to the geo graph.

    **For each** node n in cluster:

        Apply Dijkstra to get the nearest node regarding the driving distance, add it to a list. (Node 1, Node 3).

        Delete Node 1 from the iterator.

        Iterate through all nodes.

        Append {0} to the list

Iterate through all clusters

**Output**:

A route flow that represents the best route to follow from a source through all nearest node and the depo, that will be like (Node 1, Node 3, Node 2, …, {0})

Time Complexity Study:

$$\sum_{i=1}^{c}\sum_{j=1}^{n_i} o(|n_i|.T_{dk} + O(2.n_i).T_{em})$$

Where c is the number of clusters,

$n_i$ is the number of nodes in the cluster i

$T_{dk}$ is the decrease key of the data structure used for queue.

$T_{em}$ is the extract minimum of the data structure used.

### III. Conclusion:

Capacitated vehicle routing problem is an important topic that is always interpreted in the domain of optimizing, due to the wide range of use, this what make the eyes focused in this topic. In this paper I tired to implement an optimized solution that focused in the clustering part, one of the most important patterns in the routing algorithm, which is the responsible and the main effector when it comes to results, I tried to make this clustering benefits from the iterations to make it more mature.

Hence Capacitated vehicle routing problem CVRP is an NP-Hard classified problem, its hard to find the most optimal solution to solve it, all research and development tried to get near the optimal, but use cases play role.